\documentclass[10pt,twocolumn,letterpaper]{article}

\usepackage[accsupp]{axessibility}
\usepackage[pagenumbers]{wacv} % To force page numbers, e.g. for an arXiv version

\definecolor{wacvblue}{rgb}{0.21,0.49,0.74}
\usepackage[pagebackref,breaklinks,colorlinks,allcolors=wacvblue]{hyperref}
\usepackage{xcolor}
\usepackage{pifont}
\usepackage{multirow}

\def\wacvPaperID{574} % *** Enter the WACV Paper ID here
\def\confName{WACV}
\def\confYear{2027}

\title{Consommé: Conditioned Token Merging for Split Inference}

\author{Man Ngo\\
Leibniz University Hannover\\
Hannover, Germany\\
{\tt\small ngo@vss.uni-hannover.de}
\and
Jan S. Rellermeyer\\
Leibniz University Hannover\\
Hannover, Germany\\
{\tt\small rellermeyer@vss.uni-hannover.de}
}

\begin{document}
\maketitle
\begin{abstract}
Split inference partitions a Vision Transformer between a resource-constrained edge device and a server. However, transmitting intermediate tokens across the communication channel is costly. Deploying fine-tuned backbones to every endpoint becomes expensive as well. We study token reduction for split inference under a frozen backbone, where only a small adapter is shipped. We observe that the CLS token plays a dual role: it acts as the global summary the receiver needs for classification, and it serves as a signal on which patches to keep. We propose Consommé, a method utilizing both characteristics. There, an information bottleneck compresses the CLS token into a compact latent which is transmitted in place of the full CLS token and conditions a learned patch gate merging patches into exactly $k$ tokens via soft assignment. It reaches any budget in a single step, thus escaping the per-layer cap of bipartite matching (BPM). On DeiT-S/T, it leads all methods at aggressive token budgets and remains operable at shallow splits where BPM cannot operate, while matching fine-tuned methods at gentle budgets at a fraction of their deployment cost. The source code can be found here: {\url{https://github.com/Borakleon/Consomme}}.
\end{abstract}
    
\section{Introduction}

Vision Transformers (ViT) have established themselves as a staple for visual representation learning~\cite{vit}. However, their strong performance relative to CNNs comes at a cost of substantially higher computational requirements. This makes deployment on resource-constrained edge devices difficult, as the computational demands often exceed an edge device's resources~\cite{edge_overview, split_adaptation}. 

To address this, various different compression techniques have been proposed~\cite{DiffRate, tofu, Ichikawa_2026_WACV_hac-tome, bolya2022tome}. In edge computing, a main focus has been on reducing the model size by compressing it or pruning weights~\cite{wei2026nuwa, sun2025mdp, Fang_2023_CVPR, kim2026iptq, Partitioning_Edge_ICDCS}, while on less restricted devices a common solution is to reduce the number of tokens processed throughout the network by pruning and merging patch tokens~\cite{Pruning_one_more_token}.

A complementary approach partitions the network across devices. In split learning, the edge device processes the initial shallow layers on local data and transmits an intermediate representation to a more powerful device that computes the remainder~\cite{Ahuja_2023_CVPR}, leading to potential problems such as privacy and communication efficiency~\cite{split_adaptation, alvetreti2026adc, zheng2023reducing, Partitioning_Edge_ICDCS, 11044626}. Reducing the communication cost proves to be especially important, since edge devices operate under bandwidth constraints, such that the derived setting of split inference (SI) focuses solely on deployment under bandwidth and compute limits.

Efficient split inference deployment, however, is in tension with a second trend, namely, the reuse of large, publicly released checkpoints. Because the model must be present on every participating device~\cite{Partitioning_Edge_ICDCS, Ahuja_2023_CVPR}, shipping an individually fine-tuned backbone to each endpoint incurs bandwidth, storage, and energy costs. Keeping the backbone frozen and publicly shared avoids this. Providers can obtain the weights directly, deployment generalizes across devices, and a lightweight adapter adds minimal cost in exchange for task-specific behavior. This motivates token reduction that lowers transmission cost without modifying the backbone. 

Token reduction for split inference faces a tension the general efficiency literature does not: The reduced tokens must cross a bandwidth-limited channel, and the receiver still needs a global representation to classify. We observe that the classification (CLS) token is positioned to serve both needs at once. It is the global summary the receiver requires and a signal for which patches to keep. Yet, prior work uses it for only one role at a time, either as preserved intact for readout or read as an attention signal for ranking, never compressed to serve both.

Building on this, we introduce Consommé, a frozen-backbone method that compresses the CLS into a compact latent, transmits it in place of the full token, and uses it to condition a gate that merges patches into exactly k tokens by soft assignment. Because the assignment reaches any budget in a single step, the method escapes the per-layer cap of bipartite matching (BPM) while remaining feasible at aggressive budgets and shallow split points, where competing frozen methods structurally cannot produce the target count.

We summarize our contributions as follows:
\begin{itemize}
    \item A frozen-backbone soft-assignment gate reaching any budget k in one step, escaping the 50\%-per-layer cap of BPM and remaining operable at aggressive budgets and shallow splits where BPM methods cannot.
    \item A role reformulation for the CLS token, acting as a global summary and merge signal exploited by a single compressed latent that is both transmitted and used to condition merging.
    \item A characterization of the accuracy–deployment-cost trade-off across three tiers, matching fine-tuned methods at gentle budgets while shipping only a small adapter on a frozen backbone.
    \item Component ablations and split-depth analysis isolating the roles of soft-assignment, conditioning, and reconstruction.
\end{itemize}

\section{Background and Related Work}
\label{sec:Background_and_Related_Work}

\paragraph{Efficient ViTs}

\begin{table}[h]
\begin{tabular}{llll}
\hline
Tier & Method & Requires& Additional\\
&& Fine-Tuning & Parameters\\
\hline
\multirow{2}{*}{1}&ToMe~\cite{bolya2022tome}& No & No\\
&ToFu~\cite{tofu} & No & No\\
%&ATS~\cite{fayyaz2022ats} & (Yes) & No\\
\hline
\multirow{3}{*}{2}&(S/L-)HAC-ToMe~\cite{Ichikawa_2026_WACV_hac-tome}& No & Yes\\
%& S/L-HAC-ToMe~\cite{Ichikawa_2026_WACV_hac-tome}& Yes & Yes\\
&DiffRate~\cite{DiffRate} & No & Yes\\
&\textbf{Consommé (Ours)}& No & Yes\\
\hline
\multirow{2}{*}{3}&SPOT~\cite{Schlesinger_2026_CVPR_spot}& Yes & Yes\\
&DynamicViT~\cite{rao2021dynamicvit} & Yes & Yes\\
%&TokenLearner~\cite{ryoo2021tokenlearner}  & No & Yes\\
%&EViT~\cite{liang2022evit} & No & No\\
%&A-ViT~\cite{Yin_2022_CVPR_avit} & No & No\\
%&Token Pooling~\cite{TokenPooling}& No & No\\
\end{tabular}
\caption{Methods and their respective tiers. The increased footprint describes methods which have an additional overhead due to additional parameters.}
\label{tab:1_relevant}
\end{table}

In order to reduce the computational requirements of ViTs, different approaches have already been established in research. One approach focuses on reducing the number of weights within the model by pruning them~\cite{wei2026nuwa, sun2025mdp, Fang_2023_CVPR, Partitioning_Edge_ICDCS}. Another approach reduces the computational overhead of ViTs by quantizing them~\cite{kim2026iptq, 10.1007/978-3-031-72855-6_18, 10216922}.
Lastly, another procedure reduces the computational demand by reducing the number of tokens which need to be processed. This happens either by pruning or merging tokens~\cite{kong2025token}.

The approach of merging tokens is often based on BPM, introduced by ToMe~\cite{bolya2022tome} and later adapted and further optimized by DTEM~\cite{lee2024dtem} and HAC-ToMe~\cite{Ichikawa_2026_WACV_hac-tome}. BPM splits all tokens into two groups of roughly equal size and maps all tokens from one group to the most similar token in the other one, merging the $r$ most similar ones in the process, before concatenating the two sets, including the merged tokens, again. This introduces a 50\% merging limit per step. 
Reducing the number of tokens based on an attention mechanism is also often applied. A-ViT~\cite{Yin_2022_CVPR_avit}, \mbox{DynamicViT}~\cite{rao2021dynamicvit}, EViT~\cite{liang2022evit}, MCTF~\cite{lee2024mctf}, ATS~\cite{fayyaz2022ats}, SPOT~\cite{Schlesinger_2026_CVPR_spot}, Token Learner~\cite{ryoo2021tokenlearner}, and LookUpViT~\cite{LookupViT} all utilize attention in order to select which tokens to keep or to discard.
DiffRate~\cite{DiffRate} and ToFu~\cite{tofu} adapt both pruning and merging, the former via a learned schedule and the latter via a BPM adaptation.
Li~\etal~\cite{Li_2026_CVPR_Edge_RecViT} add an edge-aware token-adaptive ranking module to select tokens to select them for further computation while fine-tuning the backbone as well. Token Pooling~\cite{TokenPooling} downsamples the tokens into a target-sized subset.

The aforementioned merging and pruning methods can be divided into three different tiers. The first one centers around training- and parameter-free algorithms which can be applied off-the-shelf to frozen backbone models. In the next tier, a small trained model or a calibrated head is added to the method while still keeping the backbone frozen. The last tier harbors methods which require backbone fine-tuning while adding an additional model or calibration head as well. 
ToMe, ToFu, and ATS fall into the first tier, as they neither fine-tune the model nor add any additional parameters. HAC-ToMe, DTEM, and DiffRate fall into the second tier. All other methods fall into tier-3 since they rely on fine-tuning the backbone.
Consommé shares the tier-2 deployment profile: a frozen public backbone plus a small adapter. However, other tier-2 methods apply a BPM adaptation and therefore inherit its per-layer reduction cap. DiffRate is not capped, but its rate schedule must converge to the target, which fails at shallow splits. Consommé reaches an exact budget by construction in a single step, retaining the flexibility otherwise found only in tier-3.

\paragraph{Split Inference}

To enable the use of AI on edge devices, SI has been introduced ~\cite{split_computing_survey_2023} and received increasing attention~\cite{Partitioning_Edge_ICDCS, 11044626, 11146814, 1011453687471, 9780535}. It reduces computation on local
resource-constrained machines by splitting and distributing the workload, i.e., the computation across the model's layers.
A common goal is to find an optimal split~\cite{9780535, neurosurgeon, SPViT, DeViT}. 
To make data transfer feasible, SI also often focuses on reducing the amount of data to be transmitted by compressing the tokens~\cite{PNC, Dynamic_Compression}. This also happens by applying the previously mentioned methods, e.g. ToMe, as done by Janus~\cite{11044626}.

Edge nodes offering computation to others typically host a selection of publicly available models, which any client can use at no additional cost. Proprietary or fine-tuned models are not covered by this: they are specialized for individual use-cases and therefore of little value to a node serving many clients, so hosting them is charged as additional storage or reserved capacity. Furthermore, while a public checkpoint is already present or can be fetched by the node itself, a proprietary model must be uploaded by the client for every node it wishes to offload to, incurring bandwidth and energy costs on the constrained device. A frozen, publicly available backbone avoids both charges since only the adapter has to be supplied.
%In the spirit of edge computing and the goal of dynamic computational offloading, edge devices often host a diverse selection of free available models. The usage of proprietary or fine-tuned models, as found in tier 3, is not strictly addressed by such devices, as these models are adapted and specialized for niche use-cases which are not of interest of the public eye or rather other nodes which want to offload their computation. To offset the potential losses through hosting fewer models, a deployment of such specialized models is often charged extra. Additionally, the model still needs to be deployed there. While publicly available models can be downloaded from different sources, proprietary models need to be sent to the edge device individually. When operating in resource-constrained environments, this can have a huge impact not only on the network bandwidth but also on the available energy. 
Methods shepherded in the first tier are directly usable without the additional cost of loading individual modules into all devices. 
Second-tier methods need to load their additional learned models onto all devices which partake in the SI process. However, compared to tier-3 methods, the bandwidth and storage overhead is relatively small.

We select baselines, listed in \cref{tab:1_relevant}, to span both the deployment-cost tiers and the principal token-reduction mechanisms. From tier-1, we include ToMe and ToFu, two canonical training-free BPM methods. For ToFu, we choose to implement the MLERP version, as it achieves higher accuracy than the ToFu AVG implementation. From tier-2, we include HAC-ToMe in both variants, augmenting the BPM with attention correction and DiffRate. % which learns a per-layer compression rate. 
From tier-3, we include SPOT and DynamicViT, the most recent and general relevant work, respectively. 
SPOT utilizes the attention dynamics by taking cross-layer attention maps into account to choose the relevant tokens, while DynamicViT learns prediction masks.
Both are applied to predefined layers and are able to reach an exact token count via a keep-rate mechanism and are therefore adaptable to our setting. 

\section{Method}
\begin{figure*}[t]
    \centering
    \includegraphics[width=1.0\linewidth]{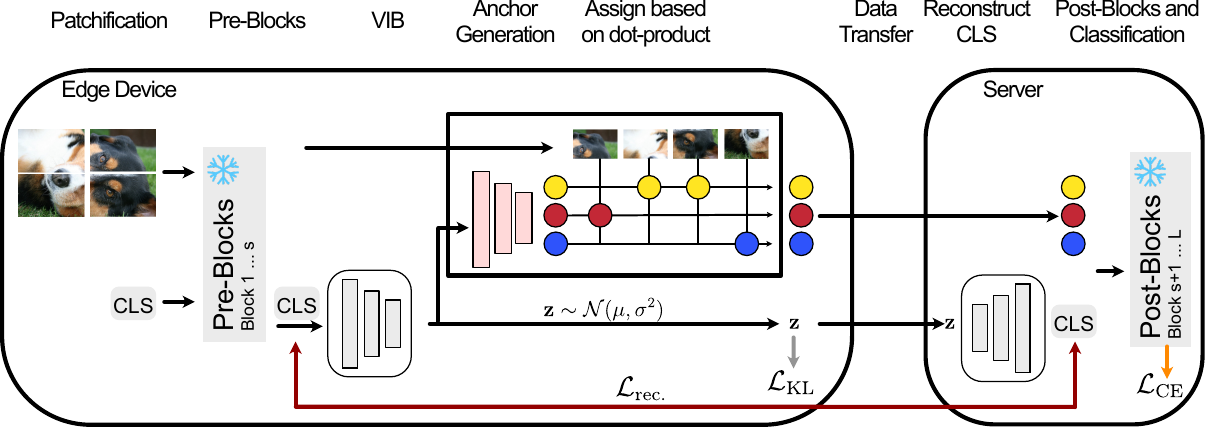}
    \caption{Overview of Consommé. By leveraging the CLS token's information via an IB, it serves as a selection mechanism for the soft-assignment merging gate in order to reduce the number of tokens to be transmitted.}
    \label{fig:method_overview}
\end{figure*}
Consommé addresses rate-constrained split inference under a frozen backbone. A pretrained ViT is partitioned at a single layer into a sender which runs the pre-split blocks on a resource-constrained edge device and a receiver which runs the post-split blocks and the classifier on an unconstrained server. The sender transmits an intermediate representation across the communication channel, from which the receiver produces the prediction. Our objective is to maximize the receiver's task accuracy subject to a transmission budget, expressed as the number of tokens crossing the channel, under the constraint that no parameter of the pretrained model is modified. Keeping the backbone frozen enables deploying the unmodified, publicly available weights while shipping only the adapter, thus avoiding the cost of hosting and distributing a fine-tuned backbone to every endpoint.

%For Consommé, we consider split inference with a frozen, pretrained ViT. A model is partitioned at a single layer into a sender performing the pre-split blocks and run on a resource-constrained edge device and a receiver running the post-split blocks and classifier on an unconstrained server. Our goal is to maximize the task accuracy subject to a transmission budget in the form of the number of tokens crossing the communication channel between sender and receiver without fine-tuning the backbone model. Keeping the backbone frozen enables the deployment of the unmodified, publicly available weights while only having to transmit the adapter, thus avoiding the cost of hosting and distributing the fine-tuned backbone to every endpoint.

%\subsection{Problem Formulation}
%We study rate-constrained split inference under a frozen backbone. Provided with a fixed, pretrained classifier partitioned into a sender and a receiver which are interconnected with a communication channel. The sender sends the receiver an intermediate representation which it then classifies by passing it through the rest of the blocks. 
%The objective is to maximize the receiver's task accuracy while minimizing the amount of information transmitted across the channel, given the constraint that no parameter of the pretrained model is to be modified. 

\subsection{Setting and Notation}
Our ViT consists of a patch embedding, a prepended CLS token, and $L$ transformer blocks followed by a classifier head. For an input, the embedding produces $N$ patch tokens, resulting in $N+1$ tokens with dimension $D$.

We partition our network at a predefined split point $s < L$. The sender then runs the blocks 1 to $s$ at full token count, only reducing it at the split point $s$. Right before the split point, it produces a CLS token $c \in \mathbb{R}^D$ and patch tokens $P \in \mathbb{R}^{N \times D}$. The receiver runs the remaining blocks starting at $s+1$ until $L$, as well as the final normalization, and the classification head. 

All backbone parameters are frozen. Only the modules, namely an information bottleneck on the CLS token and a conditioned merging gate introduced later, are trained. These produce the representations crossing the transmission channel and thus define the transmission cost: a compressed global representation $z \in \mathbb{R}^d$ with $d < D$ and $k$ merged tokens of dimensionality $D$. For each target budget $k$, we train a separate model.

SI imposes a single constraint: only the token count crossing the channel at layer $s$ matters. Progressive token reduction transmits the same payload but reaches it by removing tokens before $s$, so every subsequent pre-split block handles an already-reduced token set. Since the CLS token accumulates its global summary across depth (Sec.~\ref{ch:depth}), this degrades precisely the representation on which the reduction should be conditioned. Therefore, we retain the full token count up to $s$ and reduce once, at the split point, where reduction is actually required.
%We reduce tokens only once rather than progressively, motivated by the observation that classification-relevant information consolidates into the CLS token with depth (Sec.~\ref{ch:depth}). In early layers it remains distributed across patch tokens, so aggressive early reduction discards information later layers would have aggregated.
%, since aggressive token reduction requires representations in which classification-relevant information has been consolidated across tokens. In early representations, the information is mostly contained and distributed among all the patch tokens. Removing them aggressively also removes the information. 

\subsection{Method Overview}

At the split point, the sender compresses the CLS token $c$ into a latent representation $z$ via a variational information bottleneck. The sender uses $z$ to condition a gate which merges the $N$ patch tokens into exactly $k$ tokens. Then, it transmits the sequence $(z,M)$ with $M \in \mathbb{R}^{k\times D}$. Upon arrival, the recipient decodes $z$ into a CLS-substitute $\hat{c}$ and runs the frozen post-split blocks and classification head (Fig.~\ref{fig:method_overview}).

\subsection{CLS Information Bottleneck}
The CLS token aggregates the global image context through self-attention. We compress it through a variational information bottleneck (VIB)~\cite{vib}. The VIB serves two purposes. Firstly, it reduces the side-channel cost of transmitting global context from $D$ to $d$ values and is commonly applied for this purpose~\cite{Ahuja_2023_CVPR}. Secondly, the compressed signal serves as a compact conditioning signal $z$ for the merging gate. An encoder maps the CLS token $c$ to the parameters of the Gaussian posterior,
\begin{align}
    [\mu, \log\sigma^2] &= f_{enc}(c)\\ 
    z = \mu + \sigma \odot \epsilon,& \quad \epsilon \sim \mathcal{N}(0,I),
\end{align}
while we also sample using the reparameterization trick to enable end-to-end gradients.
Then, a decoder reconstructs the CLS substitute $\hat{\mathbf{c}} = f_{dec}(z)$ which the receiver prepends in place of the original CLS token before running the frozen post-split blocks.
%We emphasize the usage of the VIB, even though it uses the VAE's mechanism, since the VIB's objective is to retain the task-relevant global context instead of reconstructing it. The usage of a reconstruction loss is auxiliary to ensure that the decoded CLS substitute is usable by the frozen post-split blocks.

\subsection{z-Conditioned Soft-Assignment Gate}
With our global, compressed $z$ and the patch tokens $P$, the gate generates $k$ image-conditioned assignment anchors. Each patch is softly assigned across the anchors, resulting in exactly $k$ merged tokens. Because the anchors are generated from $z$, the assignment adapts to each image's global content rather than to pairwise patch similarity alone. As our ablation (Sec.~\ref{ch:ablation}) shows, the exact-budget soft-assignment is the primary source of our method's advantage, with image-conditioning providing a further, smaller gain.

%In order to generate the anchors, the gate, a small MLP, maps the patch tokens based on $z$ to $k$ low-dimensional anchor representations which are then expanded to dimension $D$ through a shared-weight matrix $W$ instead of emitting $kD$ values directly to keep the gate parameter efficient. Afterwards each patch is softly assigned over the $k$ groups by a scaled dot-product similarity normalized over each group. To obtain $k$ final tokens, each group is merged over the assignment-weighted mean. Due to the soft assignment which results in exactly $k$ output tokens for a chosen $k$, the gate remains differentiable without adding other tricks to add differentiability (e.g.~\cite{top-k}) and reaches arbitrary token budgets in a single step compared to methods relying on BPM, thus reaching aggressive $k$, even when splitting at early layers. 

A small MLP maps $z$ to $k$ low-dimensional anchors which a shared linear layer expands to dimension $D$. For each patch, we compute its dot-product similarity to all $k$ anchors with the usual attention scaling of $1/\sqrt{D}$ and apply a softmax over the anchors, resulting in a k-dimensional weight vector for each patch, representing a soft assignment so that a patch token with a comparable similarity to several anchors contributes to each of them.
%. This gives each patch a k-dimensional weight vector summing to one. A soft membership across the groups rather than a hard choice, so a patch token with comparable similarity to several anchors contributes to each of them. 
The $k$ transmitted tokens are the assignment-weighted means of the patches. 
Each anchor's token is the weighted average of all patches under the respective anchor's weights. Thus, its scale does not depend on how much total weight the anchor attracted.
The soft assignment keeps the gate differentiable without relaxation tricks (e.g. \cite{top-k}), while the $k$ anchors fix the output count, so any budget is reachable in a single step, unlike BPM, which caps reduction per layer.

\subsection{Transmission Cost}
For each image, the channel carries the compressed signal $z$ condensed into $d$ values and the $k$ merged tokens. The transmission cost can be modelled as 
\begin{align}
    cost_{\text{ours}} = (d + kD).
\end{align}
The gain of transmitting these values compared to a system with $cost_{\text{base}} = D + kD$ containing the raw signal on the side channel can be modelled as $D - d$, a fixed reduction independent of $k$. $kD$ increases with the budget whereas $D-d$ remains constant. Hence, the saving is the largest at aggressive budgets. We note that our reported comparisons exclude the CLS-token bits for all methods, so that all baselines are not charged for the full-width CLS-token they would require in a deployment, as well as the CLS-token compression.% Since a CLS-readout baseline would transmit $D$ values where we transmit $d$, excluding the bits understates our per-image saving. 

\subsection{Training}
We train only the VIB and the gate. The backbone remains frozen throughout the entire training, as per the problem statement. We optimize the cross-entropy classification loss together with the VIB. 

\begin{align}
\mathcal{L}
= \underbrace{\mathcal{L}_{\text{CE}}}_{\text{classification}} + 
\underbrace{\mathcal{L}_{\text{KL}}}_{\text{compression}} + 
 \underbrace{\mathcal{L}_\text{rec.}}_\text{reconstruction}
\end{align}

The classification loss is the primary objective. The KL term constrains the information carried by $z$ and thereby realizing the bottleneck, while the reconstruction term keeps the decoded substitute $\hat{c}$ in the distribution the frozen post-split blocks expect. Since the backbone is frozen, $c$ is a constant target.
\begin{align}
   \mathcal{L}_\text{rec.} =  \big\lVert \hat{\mathbf{c}} - \mathbf{c} \big \rVert_2^2 
\end{align}
%is applied to realize the bottleneck compression from $c$ into $\hat{c}$ while a reconstruction loss is used to aid the reconstruction of the original CLS token from the decoded substitute $\hat{c}$. The CLS token is held as a fixed target via stop-gradient $\mathrm{sg}[\cdot]$ so that $c$ can be applied as a constant target. This ensures that $\hat{c}$ or rather $c$ is usable by the frozen receiver. It is purely auxiliary. 

We add free bits to the KL~\cite{free_bits} to prevent an information bottleneck collapse. The per-dimension KL between the posterior and the standard normal prior is averaged over the batch, floored at a threshold $\tau$, and summed over the $d$ dimensions.

%It is taken between the posterior $\mathcal{N}(\mu,\sigma^2)$ and a prior. The KL terms are averaged over the whole batch to get the per-dimension value. We floor this value with a threshold $\tau$ and sum it over the dimensions $d$

Flooring removes the compression pressure from dimensions already below $\tau$, preventing individual latent dimensions from collapsing to the prior. Since the gate's output dimension scales with $k$, each budget requires its own gate. We therefore train one model per budget.

\begin{align}
    \mathcal{L}_\text{KL} &= \sum_{j=1}^d \max(\tau, \overline{\text{KL}_j}) \\
    \overline{\text{KL}_j} &= \frac{1}{B} \sum_{b=1}^B \text{KL}\left[q(z_j^{(b)} | c^{(b)})\| \mathcal{N}(0,1) \right],
\end{align}
%where $\tau$ are the free bits per dimension and $B$ is the batch size. By flooring each batch-averaged dimension KL we remove the compression pressure on dimensions which are already below $\tau$, so that the bottleneck does not collapse unused latent dimensions to the prior. 
%For each token budget we train a different model.

\begin{figure*}[t]
    \centering
    \includegraphics[width=1.0\linewidth]{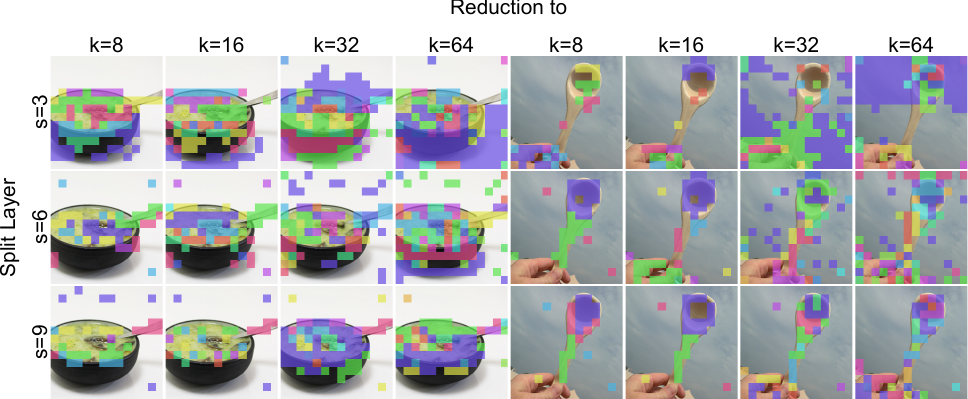}
    \caption{Consommé's visualizations of different token budgets performed on DeiT-S on different split layers $s\in\{3,6,9\}$ on samples from the ImageNet-1K validation set. The later the token reduction is performed, the more the tokens focus on the actual content and not on the background, thus identifying the more informative tokens which also aligns with the human-visual distinctive features. The biggest group which is non-colored corresponds semantically to a background group.}
    \label{fig:qualitative_results}
\end{figure*}

\begin{table}
  \centering
  \resizebox{\linewidth}{!}{
  \begin{tabular}{ll|llll}
    \multirow{2}{*}{\rotatebox[origin=c]{90}{Tier}} &Method & \multicolumn{4}{|c}{Top-1 Acc (\%)}\\
     &        & $k$=8 & $k$=16 & $k$=32 & $k$=64 \\

     \hline
     &\emph{DeiT-S} &\multicolumn{4}{|c}{79.8}\\  
    \hline

   \multirow{2}{*}{1}&ToMe&70.39$^*$&75.39$^*$&77.3$^*$&78.41$^*$\\
    &ToFu& 68.18$^*$&73.47$^*$&75.75$^*$&77.4$^*$\\    
    \hline
    \multirow{4}{*}{2}&S-HAC-ToMe& 70.83$^*$&75.72$^*$&77.54$^*$&78.56$^*$\\
    &L-HAC-ToMe&72.2$^*$&76.2$^*$&77.9$^*$&78.7$^*$\\
    &DiffRate&73.47&75.76&76.17&76.49\\
    &\textbf{Consommé}& \textbf{78.4}&\textbf{78.57}&\textbf{79.03}&79.2\\
    \hline
    \multirow{2}{*}{3}&SPOT&74.68&77.4&78.89&\textbf{79.51}\\
    &DynamicViT&    72.13&76.15&78.23&79.3\\
    \hline
     \hline
     &\emph{DeiT-T} &\multicolumn{4}{|c}{72.2}\\  
    \hline
   \multirow{2}{*}{1}&ToMe&56.66$^*$&64.56$^*$&67.78$^*$&69.66$^*$\\
    &ToFu&54.73$^*$&61.3$^*$&64.65$^*$&67.56$^*$\\    
    \hline
    \multirow{4}{*}{2}&S-HAC-ToMe& 57.94$^*$&54.31$^*$&67.91$^*$&70.13\\
    &L-HAC-ToMe&59.1$^*$&65.5$^*$&68.4$^*$&70.2$^*$\\
    &DiffRate& 62.23&66.24&68.23&69.15\\
    &\textbf{Consommé}& \textbf{69.23}&\textbf{69.91}&\textbf{70.5}&\textbf{70.95}\\
  \end{tabular}
  }
  \caption{Results of different token reduction methods on DeiT-S and DeiT-T applied to split layer 9. $^*$ indicates that the exact token count was not reached. We report the performance of the tier-3 methods only on DeiT-S as a cost-agnostic reference point given their substantially higher deployment cost for SI settings and thus omit them for DeiT-T. Consommé achieves superior results for reduced token counts.}
  \label{tab:DeiT-S/T-9}
\end{table}

\section{Experiments}
\label{ch:experiments}
We implement and evaluate Consommé based on \mbox{DeiT-S} and the smaller variant \mbox{DeiT-T}~\cite{deit}, a well-established baseline for vision transformers on ImageNet-1K~\cite{imagenet}. Further, we do the same for ViT-B~\cite{vit}, but since this does not add fundamentally new insights, we report the results in App.~\ref{sec:ViT_B_and_Depth}. We perform token reduction at layer 9 from an overall of 12. We employ a token reduction of patch tokens from 196 (excluding the CLS token) to a token transmission budget $k \in \{8,16,32,64\}$ ranging from 32\% down to 4\% of the initial token count. 
Methods utilizing several fixed token reduction layers~\cite{Schlesinger_2026_CVPR_spot, rao2021dynamicvit} are adapted to perform the token reduction at earlier, regular intervals, i.e. for split point 9, reduction points \mbox{3, 6, and 9}. Methods which reduce the token count after each block are adapted so that they only reduce the token count until the split point~\cite{bolya2022tome, Ichikawa_2026_WACV_hac-tome, tofu, DiffRate}. The token budget reduction to $k$ at the split point results in fewer tokens which need to be transmitted. This is especially useful for deployed edge devices, as they operate under limited bandwidth.
%After that layer, all adapted methods run with a constant token count to enable a fair comparison since each method runs (roughly) with the same token budget. 
Each baseline is run and, if required, fine-tuned, from their published configuration using their released training procedure and hyperparameters.
Within the classification performance evaluation of the final transmission demand, we ignore the dimensionality reduction of the CLS token, which would not be fair, as other methods do not consider this.%since Consommé also reduces the CLS token transmission, which would not be feasible in the other methods without fundamentally changing their methodology. 

\subsection{Results}
\Cref{fig:qualitative_results} shows Consommé's qualitative visualization results on ImageNet-1K samples performed on different token budgets and layers with DeiT-S. For each combination of a split layer $s\in\{3,6,9\}$ and a token budget $k\in\{8,16,32,64\}$, the image is clustered with different colors, representing the image token assignments to the $k$ different anchors. The biggest group which is non-colored corresponds semantically to a background group.
The visualizations indicate that Consommé identifies and groups tokens semantically. The groups align with semantic parts of the objects in focus. At high token budgets, the number of shown groups does not necessarily align with the token budget. This is further discussed in App.~\ref{sec:token_pruning}.
\Cref{tab:DeiT-S/T-9} reports the top-1 accuracy on DeiT-S and DeiT-T at split 9 across different token transmission budgets $k \in \{8,16,32,64\}$. 
At a more relaxed token transmission budget of $k=64$, all methods approach the performance of the base model since having more tokens available eases the token reduction pressure on the algorithm. %and allows for more potential mistakes when reducing the token count, relating to a simpler task compared to a more aggressive token reduction style.
The fine-tuned methods from tier-3 reach the highest accuracy, directly followed by Consommé. However, when reducing the token budget, Consommé outperforms all methods from all tiers on both models. Since tier-3 methods require backbone fine-tuning, we restrict them to DeiT-S as a representative upper-bound comparison rather than running them across every model/split combination.
%While the fine-tuned methods from tier 3 reach the highest accuracy in the highest token budget areas of $k=64$ with DeiT-S, Consommé achieves the highest accuracy at the lower budgets. The performance margin compared to methods in tier 2, the same tier as Consommé resides in, and the off-the-shelf methods widens as the budget tightens, thus reaching 78.4\% performance while being in a 1.4-point range of the full-token baseline (79.8\%). ToMe, the strongest tier 1 method, reaches 70.39\% and the strongest tier 2 competitor DiffRate attains 73.47\%.% However, this is still worse than SPOT, achieving a performance of 74.68\% at a token transmission budget of 8. Methods from tier 3 perform better than Consommé for a token budget of $k=64$, a token budget area where they are originally trained, fine-tuned, and optimized on, by a margin of 0.1\% and 0.31\% for DynamicViT and SPOT, respectively.

While the similarity-, importance-, and attention-based merging and pruning methods lose accuracy when forced into a smaller token budget,
Consommé performs stably, losing less than 1\% performance on \mbox{DeiT-S} and less than 2\% on \mbox{Deit-T} while reducing the token budget from $k=64$ to $k=8$. Compared to that, the other methods lose between 3-9\% on DeiT-S and 7-13\% on \mbox{DeiT-T}. 

The BPM-based methods cannot always reach the exact token budget, as they remove at most half of the remaining tokens per layer. Entries marked with $^*$ transmit the nearest reachable count rather than the stated $k$. We report the actual reached token count in Appendix~\ref{sec:ViT_B_and_Depth}. Consommé, DiffRate, SPOT, and \mbox{DynamicViT} are not subject to the reduction cap. Thus, they reach the token budget goals exactly. 

%Similarly, ~\cref{tab:DeiT-S/T-9} shows the same experiment on DeiT-T. The same trend continues there, with Consommé leading all budgets within the methods of tier 1 and tier 2, with the best competitor being ToMe from tier 1 at a token budget of $k=8$ with a difference of roughly 13\%. The consistency across the backbone scaling indicates that our model advantage is not specific to the model capacity.

\subsection{Effects of Split Depth} 
\label{ch:depth}

Token reduction is often either applied throughout the entire network or step-wise at pre-defined layers, evenly distributed throughout the network to avoid early removal of potentially informative tokens. As a consequence, the required compute decreases only in later layers.

The same principle can be applied for SI settings: the later the split point $s$ is set within the network, the more accurate information can be condensed into the CLS token, and the more layers need to be computed by the edge devices before transmission to the backend. Due to the constrained resources on the edge, it is desirable to reduce that computation as much as possible, and therefore, applying an early split is often advantageous. However, the information loss as a trade-off for compute is not to be overlooked. Hence, we analyze the impact of applying shallow split points to the network~\cite{Pruning_one_more_token}.

Instead of applying the split on layer 9, we now analyze the impact of moving the split layers to layer~6 and layer~3, respectively. Consequently, for methods applying token reduction at fixed points, we adapt them to layers 2, 4, and 6 and 1, 2, and 3, respectively. The same setting as before for methods which reduce their tokens at each layer is applied. 
\begin{figure*}
    \centering
    \includegraphics[width=1.0\linewidth]{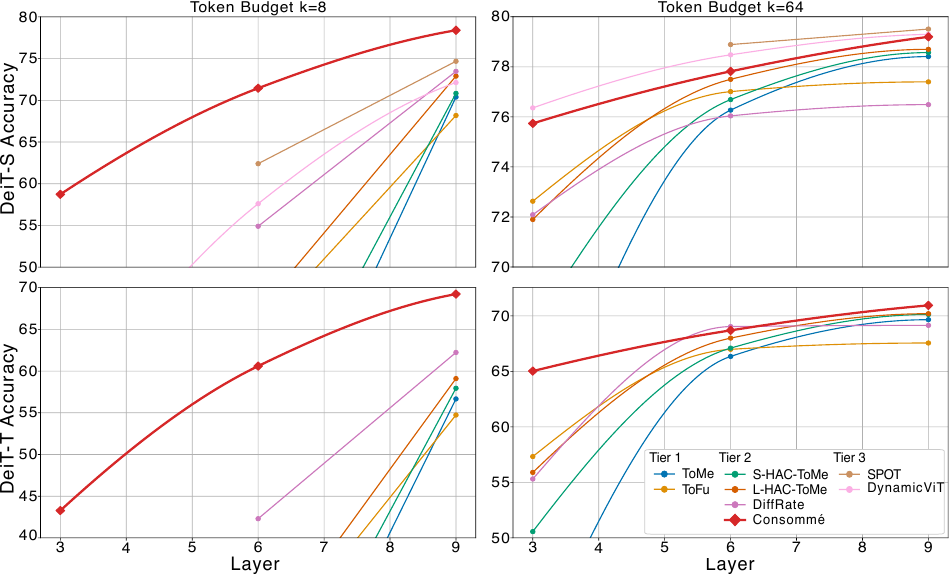}
    \caption{We evaluate the classification accuracy on DeiT-S and DeiT-T for a token budget of $k=64$ and $k=8$, while varying the split layer. Consommé exhibits much better accuracy retention at shallow splits.}
    \label{fig:interlayer_performance}
\end{figure*}

\Cref{fig:interlayer_performance} shows the results for token budgets $k=64$ and $k=8$. The performance results for $k=32$ and $k=16$ can be found in App.~\ref{sec:ViT_B_and_Depth}. While the \mbox{tier-3} methods outperform Consommé for the token budget of $k=64$, this is not the case for the strict budget. At $k=8$, Consommé performs better than the other methods. At the gentle token budget ($k=64$), most methods suffer from severe performance degradation when moving the token reduction to earlier layers on both models. %On DeiT-T, the degradation is even worse. Consommé loses about 5\% accuracy when moving from layer 9 to 3, while the other tier 1 and 2 methods lose from 10\% up to 30\% of their classification performance. 
This trend also continues on the stricter token budget $k=8$. It is noteworthy that the BPM methods cannot satisfy this budget on layer 3, as their mechanism introduces a 50\% reduction limit per reduction step. DiffRate does not converge at $s=3$ for $k=8$, while SPOT does not converge at $s=3$ at all.%DiffRate's results are also not shown, as the models do not learn a token reduction schedule reaching $k=8$ tokens. The rates converted to a different token value, thus forcing the target schedule onto the model, result in near-random accuracy, as the applied target schedule is out-of-distribution since the model was not trained on it~\cite{token_compensator}. Similarly, SPOT also could not be directly applied to split layer 3, as this resulted in a non-finite loss interrupting the training. These issues indicate a limitation of the methods under aggressive shallow-split schedules rather than an artifact of adaptation.

\subsection{Deployment and Transmission Overhead}
We distinguish two different kinds of cost which occur. Namely: a one-time per-endpoint cost of shipping parameters and storing them. %, a recurring cost factor crossing the transmission channel at inference. For split inference, both are relevant.

For deployment, as discussed in \cref{sec:Background_and_Related_Work}, the three tiers differ not only in the size of the shipped model but also in what needs to be transmitted for deployment. Tier-1 methods require nothing to be transmitted. They are applicable off-the-shelf on a frozen, publicly available backbone. Tier-2 methods retain the frozen backbone, which incurs no distribution cost, as it can be easily fetched from a public source. It adds only a small learned or calibrated set of parameters. %While DiffRate ships a set of learned per-layer compression rates, i.e. scalars, S-HAC ships head-wise calibration statistics, L-HAC contains a small learned correction network, and Consommé has an IB module and a merging gate. 
Tier-3 methods are qualitatively different as they fine-tune the backbone. The modified weights no longer match any public checkpoint, and the entire backbone must be transmitted and stored at every endpoint. %, exceeding any tier 2 overhead by an order of magnitude. This negates the premise that a device can obtain a backbone without any cost. Any predictor modules that are added, such as DynamicViT's token scoring module, are small compared to the large backbone.

%The overhead of all methods in tier 1 (none) and tier 2, except Consommé, is minimal and therefore negligible. Consommé's adapter is larger than the correction-based tier 2 methods. The reason for this is that the other methods rely on BPM as their core mechanism. The additional overhead they introduce is modules to include correction. Our adapter is substantially larger due to it not complementing an algorithm but replacing one. The information bottleneck and the conditioned gate implement a learned soft-assignment that reaches an exact budget $k$ in a single step. This mechanism enables Consommé to remain feasible at aggressive token budgets and shallow splits, where the other correction-based methods cannot produce the target token budget at all. The additional parameters are therefore the potential trade-off to function properly at such operating ranges.

The IB encoder and the gate run on the resource-constrained sender, while the IB decoder which reconstructs the CLS-substitute runs on the unconstrained receiver. The footprint on the edge device includes the encoder and the gate with 0.6M parameters, %2.7\% of DeiT-S for $k=8$ and 1.97M (9.0\%) for $k=64$, 
as seen in \cref{tab:method_overhead}. The decoder, on the other hand, remains on the server, with its size being agnostic to the token budget. Consommé adds only 10\% additional parameters in the worst configuration for a token budget of $k=64$ on DeiT-S. 

The relative parameter overhead decreases with increasing backbone width, since the backbone grows quadratically in the token dimension $D$ while the gate's budget-dependent term grows only linearly. The overhead also scales with the token budget $k$, which is expected to be low for low-power edge devices. In addition to that, the relative overhead further decreases on ViT-B, as we reduce $d$ to a third of the original CLS token size $D$~(App. \ref{sec:app_transmission_overhead}).
%While for DeiT-T, the overhead is 5\% of the original model, for ViT-B the overhead is less than 2\% being more efficient for larger models.

\begin{table}[t]
    \centering
    \begin{tabular}{lr|ccc|cc}
        Model &k&\multicolumn{3}{c|}{Additional Param (M)}& \multicolumn{2}{c}{Backbone}\\
        &  & Edge& Server& Total & & \%\\
        \hline
         DeiT-S &8 & 0.59 & 0.22 & 0.82 & 22 &3.7\\
         DeiT-S &64 & 1.97 & 0.22 & 2.2 & 22 &10\\
         DeiT-T &8 & 0.21 & 0.06 & 0.26 & 5 &5.2\\
         ViT-B & 8 & 1.1& 0.53 & 1.63 & 86&2\\
    \end{tabular}
    \caption{Consommé's deployment parameter counts on different token budgets (k) on different models compared to the backbone's parameter count.}
    \label{tab:method_overhead}
\end{table}

The sender needs to transmit the compressed global signal $z$ containing d-values together with the $k$ merged tokens per image at a cost of $d + kD$. 
The unique advantage of Consommé is its reduction of the CLS token, which saves the transmission of $D-d$ values for each inference process.
%A standard CLS baseline that also reduces the tokens to $k$ still needs to transmit a full CLS-token for classification, at a cost of $(k+1)D$. Our approach reduces the overhead by $D-d$ values per image. 

Assuming that the stored parameters of our decoder on the server and the per-image channel payload transmission are shipped at the same numerical precision as transmitted activations, Consommé's network overhead of the one-time installation on the server-side is amortized after $\sim1.1k$ images on DeiT-S compared against other token reduction algorithms. For DeiT-T, the overhead pays off at $\sim580$ images, while ViT-B repays after $\sim1k$ transmissions. 
The fine-tuned tier-3 algorithms need to ship the post-split blocks and head to the server side. When applying the split at layer $s=9$ for DeiT-S, 5.7M parameters need to be transferred, rising up to 16.4M parameters at split layer $s=3$. The transmission cost for tier-3 methods therefore grows where an early split is most attractive, whereas Consommé's does not. 
The derivation and comparison of Consommé and tier-3 algorithms to a split inference where tokens are not reduced can be found in App.~\ref{sec:app_transmission_overhead}. %We stress that the accuracy performance report (\cref{tab:DeiT-S/T-9}) excludes the CLS-token for all methods, since the baselines do not compress that token. Our advantage is therefore not reflected in the reported tables.

\subsection{Ablation Study}
\label{ch:ablation}
We conduct a series of ablation experiments to discern the functionality and the contribution of the CLS information bottleneck and the z-conditioned gate individually. All ablations use DeiT-S at split 9 and report top-1 accuracy for a token budget of $k=8$ and $k=64$.% where the receiver depends most heavily on the transmitted global signal and where each component is therefore most discriminating. We also add a token budget of $k=64$.

In order to verify the impact of the task-aware conditioning, we substitute our gating and selection mechanism with a set of statically learned, image-independent anchors while keeping and transmitting the raw CLS token, referred to as a \textbf{static anchor}. Next, within the \textbf{No-Reconstruction} version, the functionality of the decoder is inspected by using our standard gate and bottleneck method but by transmitting the original CLS token instead of $z$, thus removing the decoder entirely. This would directly mimic the mechanism of the other methods from tier-1 and 2, where these methods do not need to transmit anything to the server.
Lastly, we introduce \textbf{CLS conditioning}, removing the IB while still keeping our soft-assignment patch gate as our structural core element.

In Tab.~\ref{tab:ablation}, the no-reconstruction ablation shows reduced performance for both the aggressive $k=8$ and $k=64$ token budgets. While the difference for the high token budget remains low, it increases to 2.5\% for $k=8$. The reason for this is the loss of the reconstruction. It serves as an additional input for the IB, where the CLS information is condensed into $z$. With the removal of the reconstruction loss, the IB only relies on the classification loss which is the main task and the KL loss.

In the variation where the CLS token is directly applied to the patch gate, the IB is no longer needed, as the gate itself applies the token selection. Therefore, the method only uses the classification loss at the very end of the network to tune the patch gate. Here, the performance for the aggressive token budget performs only marginally better than the no-reconstruction approach. Once the reconstruction input is gone, it makes little difference whether the gate conditions on the compressed z or the raw CLS. Therefore, it is the reconstruction loss rather than the choice of the specific conditioning signal which accounts for the gap to the full method.

Lastly, without the usage of image-conditioned input, we achieve a performance of 74.74\%, more than 3.5\% below our full method. The static learned gate recovers most of the method's advantage over the baselines in exchange for additional parameters from the gate. Thus, we attribute the method's aggressive-budget strength primarily to the learned soft-assignment and secondarily to the image-conditioning, rather than relying solely on the conditioning.

The ablation study shows the individual contributions. The learned soft-assignment to an exact token budget provides the largest share. It keeps even the static variant above all baselines at $k=8$. The image conditioning further boosts the performance. Next, the reconstruction loss accounts for the remaining gap to the full method since it keeps the latent informative for the gate. Notably, all ablated versions outperform the baselines at $k=8$ while gradually removing single components. The CLS-conditioning and no-reconstruction ablations remove the decoder. Thus, they require nothing beyond the standard payload to be transmitted to the receiver, matching the deployment profile of tier-1 and tier-2 methods. These therefore double as lighter deployment methods.

\begin{table}[t]
    \centering
    \begin{tabular}{l|cc}
        Ablation& $k$=8& $k$=64\\
        \hline
        Consommé (full) & 78.4\phantom{0} & 79.2\phantom{0}\\
         No-Reconstruction  & 75.93 & 79.09\\
          CLS Conditioning & 76.06 & 79.02    \\
           Static Anchor  & 74.74 &  78.23   \\
    \end{tabular}
    \caption{Results of incremental component removal on DeiT-S.}
    \label{tab:ablation}
\end{table}
\section{Limitations}
Consommé assumes a CLS-readout backbone. Backbones classifying via global pooling or auxiliary token heads, such as MAE~\cite{MaskedAutoencoders2021} or LV-ViT~\cite{NEURIPS2021_lvvit}, would require adapting the global-context compression to the pooled representation.
As with all methods evaluated under the SI setting, some accuracy loss compared to the standard backbone remains even at gentle token budgets ($k=64$). Further, Consommé's cost advantage assumes a publicly hosted, shared backbone for use in public edge domains, as the usage of proprietary models is not factored into our analysis. Consommé is also less flexible in changing its token budget and split layer. There, tier 1 methods are most flexible.

\section{Conclusion}
In this paper, we present Consommé, a token reduction method for split inference. It compresses the CLS token into a compact latent, transmits it in place of the full token and uses it to condition a gate that merges the patch tokens into exactly $k$ transmitted tokens. Because the assignment reaches any budget in a single step, Consommé remains operable at aggressive budgets and shallow splits. The baselines either do not reach aggressive budgets at shallow splits or require expensive fine-tuning. Additionally, Consommé achieves higher accuracy 
with only an additional small adapter on an unmodified public checkpoint rather than a fine-tuned backbone shipped to every endpoint. We hope to see Consommé applied to on-demand edge offloading, where distributing specialized models is impractical.

%For long-running tasks, we recommend the use of our original method, while for short-term ad-hoc deployment, we encourage the usage of one of the ablated versions where the decoder does not need to be transmitted to the server. 
\newpage
{
    \small
    \bibliographystyle{ieeenat_fullname}
    \bibliography{main}

@String(IJCV = {Int. J. Comput. Vis.})

@String(CVPR= {IEEE Conf. Comput. Vis. Pattern Recog.})

@String(ECCV= {Eur. Conf. Comput. Vis.})

@String(ICLR = {Int. Conf. Learn. Represent.})

@String(IJCV  = {IJCV})

@String(CVPR  = {CVPR})

@String(ECCV  = {ECCV})

@String(ICLR  = {ICLR})

@ARTICLE{SPViT,
  author={Zhao, Sifan and Liu, Tongtong and Jin, Hai and Yao, Dezhong},
  journal={IEEE Transactions on Mobile Computing}, 
  title={SPViT: Accelerate Vision Transformer Inference on Mobile Devices via Adaptive Splitting and Offloading}, 
  year={2025},
  volume={24},
  number={10},
  pages={9303-9318},
  doi={10.1109/TMC.2025.3562721}}

@ARTICLE{DeViT,
  author={Xu, Guanyu and Hao, Zhiwei and Luo, Yong and Hu, Han and An, Jianping and Mao, Shiwen},
  journal={IEEE Transactions on Mobile Computing}, 
  title={DeViT: Decomposing Vision Transformers for Collaborative Inference in Edge Devices}, 
  year={2024},
  volume={23},
  number={5},
  pages={5917-5932},
  doi={10.1109/TMC.2023.3315138}}

@INPROCEEDINGS{PNC,
  author={Wang, Ruiqi and Liu, Hanyang and Qiu, Jiaming and Xu, Moran and Guérin, Roch and Lu, Chenyang},
  booktitle={2023 IEEE Real-Time Systems Symposium (RTSS)}, 
  title={Progressive Neural Compression for Adaptive Image Offloading Under Timing Constraints}, 
  year={2023},
  volume={},
  number={},
  pages={118-130},
  doi={10.1109/RTSS59052.2023.00020}}

@ARTICLE{Dynamic_Compression,
  author={Huang, Xiufeng and Zhou, Sheng},
  journal={IEEE Internet of Things Journal}, 
  title={Dynamic Compression Ratio Selection for Edge Inference Systems With Hard Deadlines}, 
  year={2020},
  volume={7},
  number={9},
  pages={8800-8810},
  doi={10.1109/JIOT.2020.2997128}}

@InProceedings{Ahuja_2023_CVPR,
    author    = {Ahuja, Nilesh and Datta, Parual and Kanzariya, Bhavya and Somayazulu, V. Srinivasa and Tickoo, Omesh},
    title     = {Neural Rate Estimator and Unsupervised Learning for Efficient Distributed Image Analytics in Split-DNN Models},
    booktitle = {Proceedings of the IEEE/CVF Conference on Computer Vision and Pattern Recognition (CVPR)},
    month     = {June},
    year      = {2023},
    pages     = {2022-2030}
}

@inproceedings{bolya2022tome,
  title={Token Merging: Your {ViT} but Faster},
  author={Bolya, Daniel and Fu, Cheng-Yang and Dai, Xiaoliang and Zhang, Peizhao and Feichtenhofer, Christoph and Hoffman, Judy},
  booktitle={International Conference on Learning Representations},
  year={2023}
}

@inproceedings{lee2024mctf,
  title={Multi-criteria Token Fusion with One-step-ahead Attention for Efficient Vision Transformers},
  author={Lee, Sanghyeok and Choi, Joonmyung and Kim, Hyunwoo J.},
  booktitle={Conference on Computer Vision and Pattern Recognition},
  year={2024}
}

@INPROCEEDINGS{top-k,
  author={Cordonnier, Jean-Baptiste and Mahendran, Aravindh and Dosovitskiy, Alexey and Weissenborn, Dirk and Uszkoreit, Jakob and Unterthiner, Thomas},
  booktitle={2021 IEEE/CVF Conference on Computer Vision and Pattern Recognition (CVPR)}, 
  title={Differentiable Patch Selection for Image Recognition}, 
  year={2021},
  volume={},
  number={},
  pages={2351-2360},
  doi={10.1109/CVPR46437.2021.00238}}

@inproceedings{liang2022evit,
title={Not All Patches are What You Need: Expediting Vision Transformers via Token Reorganizations},
author={Youwei Liang and Chongjian Ge and Zhan Tong and Yibing Song and Jue Wang and Pengtao Xie},
booktitle={International Conference on Learning Representations},
year={2022},
url={https://openreview.net/forum?id=BjyvwnXXVn_}
}

@InProceedings{LookupViT,
author="Koner, Rajat
and Jain, Gagan
and Jain, Prateek
and Tresp, Volker
and Paul, Sujoy",
editor="Leonardis, Ale{\v{s}}
and Ricci, Elisa
and Roth, Stefan
and Russakovsky, Olga
and Sattler, Torsten
and Varol, G{\"u}l",
title="LookupViT: Compressing Visual Information to a Limited Number of Tokens",
booktitle="Computer Vision -- ECCV 2024",
year="2025",
publisher="Springer Nature Switzerland",
address="Cham",
pages="322--337",
isbn="978-3-031-73016-0"
}

@InProceedings{Ichikawa_2026_WACV_hac-tome,
    author    = {Ichikawa, Yuki and Motomura, Masato and Van Chu, Thiem and Fujiki, Daichi},
    title     = {Efficient Vision Transformers via Token Merging with Head-wise Attention Correction},
    booktitle = {Proceedings of the IEEE/CVF Winter Conference on Applications of Computer Vision (WACV)},
    month     = {March},
    year      = {2026},
    pages     = {3908-3917}
}

@InProceedings{Schlesinger_2026_CVPR_spot,
    author    = {Schlesinger, Oded and Farzam, Amirhossein and Di Martino, J. Matias and Sapiro, Guillermo},
    title     = {SPOT: Sparsification with Attention Dynamics via Token Relevance in Vision Transformers},
    booktitle = {Proceedings of the IEEE/CVF Conference on Computer Vision and Pattern Recognition (CVPR) Findings},
    month     = {June},
    year      = {2026},
    pages     = {2335-2345}
}

@inproceedings{rao2021dynamicvit,
  title={DynamicViT: Efficient Vision Transformers with Dynamic Token Sparsification},
  author={Rao, Yongming and Zhao, Wenliang and Liu, Benlin and Lu, Jiwen and Zhou, Jie and Hsieh, Cho-Jui},
  booktitle = {Advances in Neural Information Processing Systems (NeurIPS)},
  year = {2021}
}

@article{split_computing_survey_2023,
author = {Matsubara, Yoshitomo and Levorato, Marco and Restuccia, Francesco},
title = {Split Computing and Early Exiting for Deep Learning Applications: Survey and Research Challenges},
year = {2022},
issue_date = {May 2023},
publisher = {Association for Computing Machinery},
address = {New York, NY, USA},
volume = {55},
number = {5},
issn = {0360-0300},
url = {https://doi.org/10.1145/3527155},
doi = {10.1145/3527155},
journal = {ACM Comput. Surv.},
month = dec,
articleno = {90},
numpages = {30}
}

@inproceedings{tofu,
  author={Minchul Kim and Shangqian Gao and Yen-Chang Hsu and Yilin Shen and Hongxia Jin},
  title={Token Fusion: Bridging the Gap between Token Pruning and Token Merging},
  year={2024},
  cdate={1704067200000},
  pages={1372-1381},
  url={https://doi.org/10.1109/WACV57701.2024.00141},
  booktitle={WACV},
}

@article{DiffRate,
  title={DiffRate : Differentiable Compression Rate for Efficient Vision Transformers},
  author={Mengzhao Chen and Wenqi Shao and Peng Xu, Mingbao Lin and Kaipeng Zhang and Fei Chao and Rongrong Ji and Yu Qiao and Ping Luo},
  journal={arXiv preprint arXiv:2305.17997},
  year={2023}
}

@inproceedings{NEURIPS2021_lvvit,
 author = {Jiang, Zi-Hang and Hou, Qibin and Yuan, Li and Zhou, Daquan and Shi, Yujun and Jin, Xiaojie and Wang, Anran and Feng, Jiashi},
 booktitle = {Advances in Neural Information Processing Systems},
 editor = {M. Ranzato and A. Beygelzimer and Y. Dauphin and P.S. Liang and J. Wortman Vaughan},
 pages = {18590--18602},
 publisher = {Curran Associates, Inc.},
 title = {All Tokens Matter: Token Labeling for Training Better Vision Transformers},
 url = {https://proceedings.neurips.cc/paper/2021/file/9a49a25d845a483fae4be7e341368e36-Paper.pdf},
 volume = {34},
 year = {2021}
}

@Article{MaskedAutoencoders2021,
  author  = {Kaiming He and Xinlei Chen and Saining Xie and Yanghao Li and Piotr Doll{\'a}r and Ross Girshick},
  journal = {arXiv:2111.06377},
  title   = {Masked Autoencoders Are Scalable Vision Learners},
  year    = {2021},
}

@InProceedings{ryoo2021tokenlearner,
  title={TokenLearner: Adaptive Space-Time Tokenization for Videos},
  author={Ryoo, Michael S. and Piergiovanni, AJ and Arnab, Anurag and Dehghani, Mostafa and Angelova, Anelia},
  booktitle={Advances in Neural Information Processing Systems (NeurIPS)},
  year={2021}
}

@article{fayyaz2022ats,
  title={Adaptive token sampling for efficient vision transformers},
  author={Fayyaz, Mohsen and Abbasi Kouhpayegani, Soroush and Rezaei Jafari, Farnoush 
  and Sommerlade, Eric and Vaezi Joze, Hamid Reza and Pirsiavash, Hamed and Gall, Juergen},
  journal={European Conference on Computer Vision (ECCV)},
  year={2022}
}

@InProceedings{Yin_2022_CVPR_avit,
    author    = {Yin, Hongxu and Vahdat, Arash and Alvarez, Jose M. and Mallya, Arun and Kautz, Jan and Molchanov, Pavlo},
    title     = {{A}-{V}i{T}: {A}daptive Tokens for Efficient Vision Transformer},
    booktitle = {Proceedings of the IEEE/CVF Conference on Computer Vision and Pattern Recognition (CVPR)},
    month     = {June},
    year      = {2022},
    pages     = {10809-10818}
}

@inproceedings{vit,
title={An Image is Worth 16x16 Words: Transformers for Image Recognition at Scale},
author={Alexey Dosovitskiy and Lucas Beyer and Alexander Kolesnikov and Dirk Weissenborn and Xiaohua Zhai and Thomas Unterthiner and Mostafa Dehghani and Matthias Minderer and Georg Heigold and Sylvain Gelly and Jakob Uszkoreit and Neil Houlsby},
booktitle={International Conference on Learning Representations},
year={2021},
url={https://openreview.net/forum?id=YicbFdNTTy}
}

@inproceedings{vib,
title	= {Deep Variational Information Bottleneck},author	= {Alex Alemi and Ian Fischer and Josh Dillon and Kevin Murphy},year	= {2017},URL	= {https://arxiv.org/abs/1612.00410},booktitle	= {ICLR}}

@misc{deit,
      title={Training data-efficient image transformers \& distillation through attention}, 
      author={Hugo Touvron and Matthieu Cord and Matthijs Douze and Francisco Massa and Alexandre Sablayrolles and Hervé Jégou},
      year={2021},
      eprint={2012.12877},
      archivePrefix={arXiv},
      primaryClass={cs.CV},
      url={https://arxiv.org/abs/2012.12877}, 
}

@article{imagenet,
Author = {Olga Russakovsky and Jia Deng and Hao Su and Jonathan Krause and Sanjeev Satheesh and Sean Ma and Zhiheng Huang and Andrej Karpathy and Aditya Khosla and Michael Bernstein and Alexander C. Berg and Li Fei-Fei},
Title = {{ImageNet Large Scale Visual Recognition Challenge}},
Year = {2015},
journal   = {International Journal of Computer Vision (IJCV)},
doi = {10.1007/s11263-015-0816-y},
volume={115},
number={3},
pages={211-252}
}

@inproceedings{free_bits,
 author = {Kingma, Durk P and Salimans, Tim and Jozefowicz, Rafal and Chen, Xi and Sutskever, Ilya and Welling, Max},
 booktitle = {Advances in Neural Information Processing Systems},
 editor = {D. Lee and M. Sugiyama and U. Luxburg and I. Guyon and R. Garnett},
 pages = {},
 publisher = {Curran Associates, Inc.},
 title = {Improved Variational Inference with Inverse Autoregressive Flow},
 url = {https://proceedings.neurips.cc/paper_files/paper/2016/file/ddeebdeefdb7e7e7a697e1c3e3d8ef54-Paper.pdf},
 volume = {29},
 year = {2016}
}

@INPROCEEDINGS{TokenPooling,
  author={Marin, Dmitrii and Chang, Jen-Hao Rick and Ranjan, Anurag and Prabhu, Anish and Rastegari, Mohammad and Tuzel, Oncel},
  booktitle={2023 IEEE/CVF Winter Conference on Applications of Computer Vision (WACV)}, 
  title={Token Pooling in Vision Transformers for Image Classification}, 
  year={2023},
  volume={},
  number={},
  pages={12-21},
  doi={10.1109/WACV56688.2023.00010}}

@INPROCEEDINGS {split_adaptation,
author = { Wang, Lixu and Shang, Bingqi and Li, Yi and Mohapatra, Payal and Dong, Wei and Wang, Xiao and Zhu, Qi },
booktitle = { 2025 IEEE/CVF Conference on Computer Vision and Pattern Recognition (CVPR) },
title = {{ Split Adaptation for Pre-trained Vision Transformers }},
year = {2025},
volume = {},
ISSN = {},
pages = {20092-20102},
doi = {10.1109/CVPR52734.2025.01871},
url = {https://doi.ieeecomputersociety.org/10.1109/CVPR52734.2025.01871},
publisher = {IEEE Computer Society},
address = {Los Alamitos, CA, USA},
month =Jun}

@ARTICLE{edge_overview,
  author={Cao, Keyan and Liu, Yefan and Meng, Gongjie and Sun, Qimeng},
  journal={IEEE Access}, 
  title={An Overview on Edge Computing Research}, 
  year={2020},
  volume={8},
  number={},
  pages={85714-85728},
  doi={10.1109/ACCESS.2020.2991734}}

@inproceedings{wei2026nuwa,
  title={NuWa: Deriving Lightweight Class-Specific Vision Transformers for Edge Devices},
  author={Wei, Ziteng and He, Qiang and Li, Bing and Chen, Feifei and Jin, Hai and Yang, Yun},
  booktitle={Proceedings of the IEEE/CVF Conference on Computer Vision and Pattern Recognition},
  pages={156--166},
  year={2026}
}

@inproceedings{sun2025mdp,
  title={MDP: multidimensional vision model pruning with latency constraint},
  author={Sun, Xinglong and Lakshmanan, Barath and Shen, Maying and Lan, Shiyi and Chen, Jingde and Alvarez, Jose M},
  booktitle={Proceedings of the Computer Vision and Pattern Recognition Conference},
  pages={20113--20123},
  year={2025}
}

@InProceedings{Fang_2023_CVPR,
    author    = {Fang, Gongfan and Ma, Xinyin and Song, Mingli and Mi, Michael Bi and Wang, Xinchao},
    title     = {DepGraph: Towards Any Structural Pruning},
    booktitle = {Proceedings of the IEEE/CVF Conference on Computer Vision and Pattern Recognition (CVPR)},
    month     = {June},
    year      = {2023},
    pages     = {16091-16101}
}

@inproceedings{kim2026iptq,
  title={Iptq-vit: Post-training quantization of non-linear functions for integer-only vision transformers},
  author={Kim, Gihwan and Lee, Jemin and Kim, Hyungshin},
  booktitle={Proceedings of the IEEE/CVF Winter Conference on Applications of Computer Vision},
  pages={7366--7375},
  year={2026}
}

@inproceedings{alvetreti2026adc,
  title     = {Communication Efficient Split Learning of {ViTs} with
               Attention-based Double Compression},
  author    = {Alvetreti, Federico and Pomponi, Jary and
               Di Lorenzo, Paolo and Scardapane, Simone},
  booktitle = {Proceedings of the 34th European Signal Processing Conference
               (EUSIPCO)},
  year      = {2026},
}

@INPROCEEDINGS {Pruning_one_more_token,
author = { Eliopoulos, Nicholas John and Jajal, Purvish and Davis, James C. and Liu, Gaowen and Thiravathukal, George K. and Lu, Yung-Hsiang },
booktitle = { 2025 IEEE/CVF Winter Conference on Applications of Computer Vision (WACV) },
title = {{ Pruning One More Token is Enough: Leveraging Latency-Workload Non-Linearities for Vision Transformers on the Edge }},
year = {2025},
volume = {},
ISSN = {},
pages = {7153-7162},
doi = {10.1109/WACV61041.2025.00695},
url = {https://doi.ieeecomputersociety.org/10.1109/WACV61041.2025.00695},
publisher = {IEEE Computer Society},
address = {Los Alamitos, CA, USA},
month =mar}

@article{zheng2023reducing,
  title={Reducing communication for split learning by randomized top-k sparsification},
  author={Zheng, Fei and Chen, Chaochao and Lyu, Lingjuan and Yao, Binhui},
  journal={arXiv preprint arXiv:2305.18469},
  year={2023}
}

@InProceedings{Li_2026_CVPR_Edge_RecViT,
    author    = {Li, YiZhou and Xu, Jinyi and Yin, Mingyu and Zhao, Xianyi},
    title     = {Edge-RecViT: Efficient Vision Transformer via Semantic-Refined Dynamic Recursion},
    booktitle = {Proceedings of the IEEE/CVF Conference on Computer Vision and Pattern Recognition (CVPR)},
    month     = {June},
    year      = {2026},
    pages     = {12987-12996}
}

@INPROCEEDINGS{Partitioning_Edge_ICDCS,
  author={Liu, Xiang and Song, Yijun and Li, Xia and Sun, Yifei and Lan, Huiying and Liu, Zemin and Jiang, Linshan and Li, Jialin},
  booktitle={2025 IEEE 45th International Conference on Distributed Computing Systems (ICDCS)}, 
  title={Efficient Partitioning Vision Transformer on Edge Devices for Distributed Inference}, 
  year={2025},
  volume={},
  number={},
  pages={286-296},
  doi={10.1109/ICDCS63083.2025.00036}}

@article{neurosurgeon,
author = {Kang, Yiping and Hauswald, Johann and Gao, Cao and Rovinski, Austin and Mudge, Trevor and Mars, Jason and Tang, Lingjia},
title = {Neurosurgeon: Collaborative Intelligence Between the Cloud and Mobile Edge},
year = {2017},
issue_date = {April 2017},
publisher = {Association for Computing Machinery},
address = {New York, NY, USA},
volume = {52},
number = {4},
issn = {0362-1340},
url = {https://doi.org/10.1145/3093336.3037698},
doi = {10.1145/3093336.3037698},
journal = {SIGPLAN Not.},
month = apr,
pages = {615–629},
numpages = {15}
}

@inproceedings{10.1007/978-3-031-72855-6_18,
author = {Ramachandran, Akshat and Kundu, Souvik and Krishna, Tushar},
title = {CLAMP-ViT: Contrastive Data-Free Learning for Adaptive Post-training Quantization of ViTs},
year = {2024},
isbn = {978-3-031-72854-9},
publisher = {Springer-Verlag},
address = {Berlin, Heidelberg},
url = {https://doi.org/10.1007/978-3-031-72855-6_18},
doi = {10.1007/978-3-031-72855-6_18},
booktitle = {Computer Vision – ECCV 2024: 18th European Conference, Milan, Italy, September 29–October 4, 2024, Proceedings, Part LXVII},
pages = {307–325},
numpages = {19},
location = {Milan, Italy}
}

@ARTICLE{10216922,
  author={Li, Zhikai and Chen, Mengjuan and Xiao, Junrui and Gu, Qingyi},
  journal={IEEE Transactions on Neural Networks and Learning Systems}, 
  title={PSAQ-ViT V2: Toward Accurate and General Data-Free Quantization for Vision Transformers}, 
  year={2024},
  volume={35},
  number={12},
  pages={17227-17238},
  doi={10.1109/TNNLS.2023.3301007}}

@inproceedings{
    lee2024dtem,
    title={Learning to Merge Tokens via Decoupled Embedding for Efficient Vision Transformers},
    author={Dong Hoon Lee and Seunghoon Hong},
    booktitle={Conference on Neural Information Processing Systems},
    year={2024},
}

@article{kong2025token,
  title={Token Reduction Should Go Beyond Efficiency in Generative Models--From Vision, Language to Multimodality},
  author={Kong, Zhenglun and Li, Yize and Zeng, Fanhu and Xin, Lei and Messica, Shvat and Lin, Xue and Zhao, Pu and Kellis, Manolis and Tang, Hao and Zitnik, Marinka},
  journal={arXiv preprint arXiv:2505.18227},
  year={2025}
}

@ARTICLE{9780535,
  author={Tuli, Shreshth and Casale, Giuliano and Jennings, Nicholas R.},
  journal={IEEE Transactions on Mobile Computing}, 
  title={SplitPlace: AI Augmented Splitting and Placement of Large-Scale Neural Networks in Mobile Edge Environments}, 
  year={2023},
  volume={22},
  number={9},
  pages={5539-5554},
  doi={10.1109/TMC.2022.3177569}}

@INPROCEEDINGS{11044626,
  author={Jiang, Linyi and Fu, Silvery D. and Zhu, Yifei and Li, Bo},
  booktitle={IEEE INFOCOM 2025 - IEEE Conference on Computer Communications}, 
  title={Janus: Collaborative Vision Transformer Under Dynamic Network Environment}, 
  year={2025},
  volume={},
  number={},
  pages={1-10},
  doi={10.1109/INFOCOM55648.2025.11044626}}

@ARTICLE{11146814,
  author={Xu, Guanyu and Hao, Zhiwei and Shen, Li and Luo, Yong and Sun, Fuhui and Wang, Xiaoyan and Hu, Han and Wen, Yonggang},
  journal={IEEE Transactions on Computers}, 
  title={CoFormer: Collaborating With Heterogeneous Edge Devices for Scalable Transformer Inference}, 
  year={2025},
  volume={74},
  number={12},
  pages={4010-4024},
  doi={10.1109/TC.2025.3604473}}

@article{1011453687471,
author = {Mudvari, Akrit and Vainio, Antero and Ofeidis, Iason and Tarkoma, Sasu and Tassiulas, Leandros},
title = {Adaptive Compression-Aware Split Learning and Inference for Enhanced Network Efficiency},
year = {2024},
issue_date = {November 2024},
publisher = {Association for Computing Machinery},
address = {New York, NY, USA},
volume = {24},
number = {4},
issn = {1533-5399},
url = {https://doi.org/10.1145/3687471},
doi = {10.1145/3687471},
journal = {ACM Trans. Internet Technol.},
month = nov,
articleno = {27},
numpages = {26}
}
}
\newpage

\maketitlesupplementary
\section{Results}
\label{sec:ViT_B_and_Depth}
\begin{table}[t]
\resizebox{\linewidth}{!}{%
  \centering
  \begin{tabular}{ll|llll}
    \multirow{2}{*}{\rotatebox[origin=c]{90}{Tier}}&Method & \multicolumn{4}{|c}{Top-1 Acc (\%)}\\
   &         & $k$=8 & $k$=16 & $k$=32 & $k$=64  \\
    \hline
     &\emph{ViT-B} &\multicolumn{4}{|c}{85.1}\\
    \hline
   \multirow{2}{*}{1}&ToMe&60.88$^*$&75.98$^*$&80.7$^*$&83.03$^*$\\
    &ToFu&70.96$^*$&78.7$^*$&81.79$^*$&83.18$^*$\\    
    \hline
    \multirow{4}{*}{2}&S-HAC-ToMe& 65.1$^*$&77.23$^*$&81.41$^*$&83.22\\
    &L-HAC-ToMe&69.81$^*$&78.61$^*$&82.07$^*$&83.7$^*$\\
    &DiffRate& 70.75&74.49&75.89&78.38\\
    &\textbf{Consommé}& \textbf{82.58}&\textbf{83.4}&\textbf{83.87}&\textbf{84.16}\\
  \end{tabular}
  }
  \caption{Results of different token reduction methods on ViT-B applied to split layer $s=9$. $^*$ indicates that the exact token count was not reached. Consommé achieves superior results on all reported token budgets $k$.}
  \label{tab:ViT-B-9}
\end{table}

\begin{table}
    \centering
    \begin{tabular}{c|cccc}
        \multirow{2}{*}{Split $s$} & \multicolumn{4}{c}{Transmitted patch tokens} \\
        & $k$=8 & $k$=16 & $k$=32 & $k$=64 \\
        \hline
        3 & - & - & 32 & 64 \\
        6 & \phantom{0}8 & 16 & \textbf{34} & 64 \\
        9 & \phantom{0}\textbf{9} & \textbf{18} & \textbf{34} & \textbf{61} \\
    \end{tabular}
    \caption{Number of patch tokens actually transmitted by the BPM-based methods (ToMe, ToFu, S-HAC, L-HAC). The CLS token is not counted. Because bipartite matching removes at most half the tokens per layer, the reachable count is higher than the token budget $k$ on $s=3$. We report the nearest reachable count. - marks budgets that are unreachable at that split.}
    \label{tab:transmitted_tokens}
\end{table}

\begin{table}[t]
  \centering
  \resizebox{\linewidth}{!}{%

  \begin{tabular}{ll|llll}
    \multirow{2}{*}{\rotatebox[origin=c]{90}{Tier}}&Method & \multicolumn{4}{|c}{Top-1 Acc (\%)}\\
     &        & $k$=8 & $k$=16 & $k$=32 & $k$=64 \\

    \hline
     &\emph{DeiT-S} &\multicolumn{4}{|c}{79.8}\\  
    \hline
   \multirow{2}{*}{1}&ToMe&19.63&53.22&70.72$^*$&76.27\\
    &ToFu& 42.39&62.49&73.27$^*$&77.01\\    
    \hline
    \multirow{4}{*}{2}&S-HAC-ToMe& 26.42&57.4&71.95$^*$&76.69\\
    &L-HAC-ToMe&44.7&63.3&73.9$^*$&77.5\\
    &DiffRate&  54.9&69.57&74.8&76.04\\
    &\textbf{Consommé}& \textbf{71.45}&\textbf{74.78}&76.64&77.82\\
    \hline
    \multirow{2}{*}{3}&SPOT&62.4&72.4&\textbf{76.98}&\textbf{78.89}\\
    &DynamicViT&    57.6&69.97&75.93&78.48\\
     \hline
    \hline
     &\emph{DeiT-T} &\multicolumn{4}{|c}{72.2}\\
    \hline
   \multirow{2}{*}{1}&ToMe&\phantom{0}8.63$^*$&35.6$^*$&57.05$^*$&66.35\\
    &ToFu& 25.1&46.85&61.14$^*$&66.99\\    
    \hline
    \multirow{4}{*}{2}&S-HAC-ToMe& 13.74&41.56&59.36$^*$&67.1\\
    &L-HAC-ToMe&26.5&48.8&62.1$^*$&68.0\\
    &DiffRate&  42.3&57.84&66.05&\textbf{69.04}\\
    &\textbf{Consommé} & \textbf{60.6}&\textbf{64.46}&\textbf{67.2}&68.7\\

     \hline
    \hline
     &\emph{ViT-B} &\multicolumn{4}{|c}{85.1}\\
    \hline
   \multirow{2}{*}{1}&ToMe&\phantom{0}4.17$^*$&29.19$^*$&66.93$^*$&79.9\\
    &ToFu& 23.17&57.23&76.4$^*$&82.12\\    
    \hline
    \multirow{4}{*}{2}&S-HAC-ToMe& 10.21&43.56&71.29$^*$&80.85\\
    &L-HAC-ToMe&36.17&61.3&76.4$^*$&81.9\\
    &DiffRate&  26.6&56.98&72.55&77.22\\
    &\textbf{Consommé} & \textbf{73.87}&\textbf{78.0}&\textbf{80.7}&\textbf{82.74}\\
  \end{tabular}
  }
  \caption{Split 6 results of different token reduction methods on DeiT-S, DeiT-T, and ViT-B. $^*$ indicates that the exact token count was not reached by the method. On all models, methods from all three tiers suffer from heavy performance degradation, with some even reaching single-digit top-1 accuracy. Consommé achieves superior results when reducing the token budget.}  
  \label{tab:split_6_all}
\end{table}

\begin{table}[t]
  \centering
  \begin{tabular}{ll|llll}
    \multirow{2}{*}{\rotatebox[origin=c]{90}{Tier}}&Method & \multicolumn{4}{|c}{Top-1 Acc (\%)}\\
     &         & $k$=8 & $k$=16 & $k$=32 & $k$=64 \\

     \hline
     &\emph{DeiT-S}&\multicolumn{4}{|c}{79.8}\\  
    \hline
   \multirow{2}{*}{1}&ToMe&-&-&19.1&61.57\\
    &ToFu&-&-&52.18&72.63\\    
    \hline
    \multirow{4}{*}{2}&S-HAC-ToMe&-&-&35.58&67.54\\
    &L-HAC-ToMe&-&-&51.67&71.9\\
    &DiffRate&$\circ$&$\circ$&49.5&72.09\\
    &\textbf{Consommé}& \textbf{58.73}&\textbf{65.9}&\textbf{70.99}&75.74\\
    \hline
    \multirow{2}{*}{3}&SPOT  &$\circ$&$\circ$&$\circ$&$\circ$\\
    &DynamicViT&    30.76&52.47&68.7&\textbf{76.36}\\

     \hline
    \hline
     &\emph{DeiT-T}  &\multicolumn{4}{|c}{72.2}\\  
    \hline
   \multirow{2}{*}{1}&ToMe&-&-&\phantom{0}6.26&39.04\\
    &ToFu&-&-&30.23&57.33\\    
    \hline
    \multirow{4}{*}{2}&S-HAC-ToMe&-&-&18.09&50.57\\
    &L-HAC-ToMe&-&-&32.0&55.9\\
    &DiffRate& $\circ$ &$\circ$&24.8&55.31\\
    &\textbf{Consommé}& \textbf{43.28}&\textbf{51.12}&\textbf{58.59}&\textbf{65.03}\\

     \hline
    \hline
     &\emph{ViT-B}&\multicolumn{4}{|c}{85.1}\\
    \hline
   \multirow{2}{*}{1}&ToMe&-&-&10.68&65.07\\
    &ToFu&-&-&53.97&77.86\\    
    \hline
    \multirow{4}{*}{2}&S-HAC-ToMe&-&-&31.98&73.03\\
    &L-HAC-ToMe&-&-&58.8&77.9\\
    &DiffRate& $\circ$ &$\circ$&31.68&69.77\\
    &\textbf{Consommé} & \textbf{58.89}&\textbf{68.14}&\textbf{75.45}&\textbf{80.41}\\
  \end{tabular}
  \caption{Results on split layer 3. $\circ$ marks that the model did not converge. - shows that the token count was not reached by the method. Consommé satisfies all token budgets $k$ on split layer $s=3$, whereas all BPM methods, as well as some training-based methods, fail.}
  \label{tab:split_3_all}
\end{table}
\Cref{tab:ViT-B-9} reports the SI classification accuracy of the tier-1 and tier-2 methods on ViT-B on split layer $s=9$ for token transmission budgets $k\in\{8,16,32,64\}$. Similar to the experiments on DeiT-S and DeiT-T~(\cref{tab:DeiT-S/T-9}), Consommé outperforms the other tier-1 and tier-2 methods. Moreover, when reducing the token budget from $k=64$ to $k=8$, Consommé exhibits higher performance retention compared to the other methods. While ViT-B in general achieves higher top-1 accuracy on ImageNet-1K and most methods achieve higher top-1 accuracy on ViT-B compared to their performance on DeiT-S at token budget $k=64$, many perform worse than on DeiT-S at token budget $k=8$. The performance degradation of the other methods on ViT-B spans roughly from 8\% up to 23\%, which is even higher than the performance degradation on DeiT-T. 

\subsection{Effects of Split Depth}
\Cref{tab:transmitted_tokens} reports the transmitted tokens for the BPM methods on the different layers $s$ for the corresponding token budgets $k$. We report the nearest feasible token budget if the direct token budget cannot be reached. On all $s=9$ and $s=6$ $k=32$ configurations, the token count differs by at most 3 tokens from the budget limit. The token budget on $s=9$ at $k=64$ is the only configuration where the reached token count is below the token limit $k$. Except for $s=3$ on $k=8$ and $k=16$, all other token limits are completely satisfied. There, the BPM reach a token count of 25, differing by more than 9 tokens from the budget limit.

\Cref{tab:split_6_all} and \Cref{tab:split_3_all} provide the exact values of the analysis of the split layers 3 and 6 discussed in \cref{ch:depth}. The accuracy falls for all methods as the split moves to earlier layers, and Consommé's performance margin over the other methods widens. On $s=3$ the BPM-based methods cannot reach the target due to their per-step reduction limit. DiffRate's rate schedule does not converge to the target, and SPOT's training loss becomes non-finite.

\subsection{Deployment and Transmission Overhead}
\label{sec:app_transmission_overhead}
\begin{table*}[t]
    \centering
    \begin{tabular}{l|rrr}
        & DeiT-T & DeiT-S & ViT-B \\
        Token dimension $D$ / latent $d$ & 192 / 96 & 384 / 192 & 768 / 256 \\
        \hline
        \emph{Per image (values)} & & & \\
        \quad Full activations $(N+1)D$ & 37,824 & 75,648 & 151,296 \\
        \quad Consommé $d+kD$             &  1,632 &  3,264 &   6,400 \\
        \quad Compressed Baseline Method $(1+k)D$             &  1,782 &  3,456 &   6,912 \\
        %\quad saving                      & 36,192 & 72,384 & 144,896 \\
        %\quad reduction                   & 95.7\% & 95.7\% &  95.8\% \\
        \hline
        \emph{Consommé deployment (parameters)} & & & \\
        %\quad edge (encoder + gate)       &   205k &   593k &   1.10M \\
        \quad server (decoder)            &    56k &   223k &    527k \\
        \quad total                       &   261k &   816k &   1.63M \\
        \hline
        \emph{Tier-3 deployment (parameters, server only)} & & & \\
        \quad $s=9$                       &  1.53M &  5.71M &   22.0M \\
        \quad $s=6$                       &  2.86M &  11.0M &   43.3M \\
        \quad $s=3$                       &  4.20M &  16.4M &   64.6M \\
        \hline
        \emph{Break-even vs.\ compressed baseline (images)} & & & \\
        \quad Consommé, server only       &    584 &  1,160 &   1,028 \\
        \hline
        \emph{Amortization vs.\ full-activation split (images)} & & & \\
        \quad Consommé, total adapter     &    7.2 &   11.3 &    11.2 \\
        \quad Consommé, server only       &    1.6 &    3.1 &     3.6 \\
        \quad tier-3 server, $s=9$        &     42 &     79 &     153 \\
        \quad tier-3 server, $s=6$        &     79 &    153 &     300 \\
        \quad tier-3 server, $s=3$        &    116 &    227 &     447 \\
    \end{tabular}    
    \caption{Transmission and deployment cost at token budget $k=8$. The Break-even point is the one-time shipping cost divided by the per-image saving against the stated reference, assuming that parameters and activations are shipped at equal numerical precision. Consommé's decoder size and cost are split-layer invariant, whereas the tier-3 cost grows as the split layer moves earlier.}
    \label{tab:method_amortization}
\end{table*}

\Cref{tab:method_amortization} shows the different parameter and transmission costs of sending all tokens, the tier-3 reduced activations at a token budget $k=8$, and Consommé's sent activations at $k=8$.

Consommé's CLS token reduction to latent $d$ is set to be at most 50\% of the original token size. We set $d$ to 50\% for DeiT-T and DeiT-S. For ViT-B, we reduce this further to d=256 to limit the additional growth in adapter size that a larger backbone would otherwise entail.
When compared to a CLS-readout baseline, as a tier-1 method, both approaches reduce the token count to $k$ which is sent over to the server in addition to the CLS token, resulting in the break-even point for Consommé of 584, 1.2k, and 1k images to offset the cost of the shipped decoder.

When comparing Consommé and a tier-3 method to the full-activation split, it needs to be considered that Consommé's shipped decoder is size-invariant. Thus, the number of parameters which need to be sent to the server remains the same, regardless of the split layer $s$. However, this is not the case for a tier-3 method, as it needs to transfer the fine-tuned model weights for the corresponding layers. If the split layer is applied for $s=3$, the remaining 9 layers need to be transmitted to the server for our model selection. This cost is amortized after doing SI for more than 100 images on DeiT-T, while Consommé's break-even point is reached after at most 8 images on DeiT-T.

\subsection{Anchor Assignments and Pruning}
\label{sec:token_pruning}
As shown in \cref{fig:qualitative_results}, the number of used anchors sometimes differs from the token budget $k$. This is due to Consommé's soft-assignment mechanism. Each patch is assigned across the $k$ anchors by a softmax. Therefore, each token can contribute to several anchors at once. Summing a group's weights gives its occupancy in units of patches, the equivalent number of patches that contributed to it. A group with a weight of 3.0 consists of an equivalent of three patch tokens. Since each patch token has a total weight of 1, the total weight is $N$, and the mean per anchor is $N/k$, resulting in 24.5 patches at $k=8$ and 3.06 for $k=64$.
The qualitative results in ~\cref{fig:qualitative_results} show that this mass is not distributed equally throughout the anchors. Large homogeneous regions collapse into a single group (e.g. corresponding to the background), while other anchors receive very little weight. We quantify this distribution by introducing a threshold $\tau$. Every anchor group with a total weight below $\tau$ patches is pruned and therefore not transmitted. We report the classification accuracy, as well as the final transmitted token distribution in \cref{tab:token_pruning}. For a token budget of $k=64$, removing anchors with less than 0.1 token weight leads to 38.3 tokens transmitted on average for a performance reduction of 0.31\%. Increasing the threshold $\tau$ leads to further performance degradation for $k=64$, while $k=8$ performs stably. For $k=64$, many anchors carry little token weight. 
The assignment is image-dependent, resulting in the number of transmitted tokens varying heavily when pruning. With additional pruning, $k$ acts as an upper bound for the token budget. 
The low-mass anchor groups are nonetheless informative rather than noise, as removing them reduces the model's performance. Pruning therefore is also not a substitute for the adapter at a different target budget. Pruning a $k=64$ model to roughly $k=16$ tokens reaches 77.15\%. A model directly on this token budget achieves 78.57\%. We therefore report the exact-k configuration throughout. Pruning low-weight anchors offers a further rate-accuracy trade-off where a variable payload is acceptable, and no model for the lower budget is available.

\begin{table}[t]
    \centering
    \begin{tabular}{cc|cccc}
        &&&\multicolumn{3}{c}{Tokens Kept}\\
        $k$ & $\tau$ & Acc (\%) & Mean & Min & Max \\
        \hline
        \multirow{6}{*}{8}
           & 0.0 & 78.40 & $8.00\pm0.00$ & 8 & 8 \\
           & 0.1 & 78.40 & $7.97\pm0.19$ & 5 & 8 \\
           & 0.5 & 78.40 & $7.77\pm0.58$ & 3 & 8 \\
           & 1.0 & 78.36 & $7.43\pm0.91$ & 2 & 8 \\
           & 2.0 & 78.03 & $6.62\pm1.29$ & 2 & 8 \\
           & 3.0 & 77.02 & $5.80\pm1.47$ & 1 & 8 \\
        \hline
        \multirow{6}{*}{64}
           & 0.0 & 79.20 & $64.0\pm0.0$ & 64 & 64 \\
           & 0.1 & 78.89 & $38.3\pm7.5$  & 8 & 60 \\
           & 0.5 & 78.30 & $23.3\pm6.1$  & 5 & 46 \\
           & 1.0 & 77.15 & $16.1\pm4.6$  & 3 & 33 \\
           & 2.0 & 74.95 & $\;9.5\pm2.9$ & 1 & 22 \\
           & 3.0 & 73.01 & $\;6.8\pm2.0$ & 1 & 17 \\
    \end{tabular}
    \caption{Token pruning on DeiT-S at split $s=9$. Anchor groups with a summed assignment weight less than $\tau$ tokens are pruned and not transmitted.}
    \label{tab:token_pruning}
\end{table}

\end{document}